\def\year{2020}\relax
\documentclass[letterpaper]{article} 
\usepackage{aaai20}  
\usepackage{times}  
\usepackage{helvet} 
\usepackage{courier}  
\usepackage[hyphens]{url}  
\usepackage{graphicx} 
\usepackage{graphicx}  

\usepackage{multirow}
\usepackage{booktabs}

\usepackage{amsfonts}
\usepackage{amsmath}

\title{HySTER: A Hybrid Spatio-Temporal Event Reasoner}
\author{\Large \textbf{Theophile Sautory, Nuri Cingillioglu, Alessandra Russo}\\ 
Imperial College London, Department of Computing\\ 
London SW7 2BU, United Kingdom\\
\{theophile.sautory15, nuric, a.russo\}@imperial.ac.uk 
}
\begin{document}

\maketitle

\begin{abstract}
The task of Video Question Answering (VideoQA) consists in answering natural language questions about a video and serves as a proxy to evaluate the performance of a model in scene sequence understanding. Most methods designed for VideoQA up-to-date are end-to-end deep learning architectures which struggle at complex temporal and causal reasoning and provide limited transparency in reasoning steps. We present the HySTER: a Hybrid Spatio-Temporal Event Reasoner to reason over physical events in videos. Our model leverages the strength of deep learning methods to extract information from video frames with the reasoning capabilities and explainability of symbolic artificial intelligence in an answer set programming framework. We define a method based on general temporal, causal and physics rules which can be transferred across tasks. We apply our model to the CLEVRER dataset and demonstrate state-of-the-art results in question answering accuracy. This work sets the foundations for the incorporation of inductive logic programming in the field of VideoQA.
\end{abstract}

\section{Introduction}
\noindent Visual reasoning over videos is a task relevant for autonomous agents performing in real world applications such as autonomous vehicles or dynamic robot manipulation. The task of Video Question Answering (VideoQA) consists of the following: given a video alongside a question in natural language a model must return an answer in natural language \cite{video_qa}. Due to the joint reasoning on vision, language, and spatio-temporal properties, VideoQA serves to evaluate visual reasoning over videos.

In recent years, two main approaches to solve VideoQA and Visual Question Answering (VQA, the equivalent on single images) tasks have been explored. \cite{mario_qa,vqa_gradually_ref_attention} rely on attention mechanisms and fully end-to-end differentiable systems to extract the relevant information from the image frames to answer questions. Their reasoning steps and failure mechanisms remain opaque to humans whilst requiring large amounts of data to train.

\cite{n2nmn,MAC,ns_vqa,clevrer} employ neuro-symbolic methods which perform compositional reasoning by breaking down the answering task into applying a series of symbolic filters to extract the answer. Neural network filters \cite{n2nmn,MAC} allow the interpretation of respective attention feature maps for each reasoning step, but do not explain the decision making process of the model and still consume thousands of samples to train. Heuristic filters with functional programs have been used \cite{ns_vqa,clevrer} and elucidate the decision process of the model. However, they do not benefit from the full representational power of symbolic reasoning methods such as answer set programming (ASP) and  would have difficulties generalising to new symbolic domains.

Motivated by the potential of neuro-symbolic methods to simulate human intelligence and the application of Kahneman’s human’s thinking systems: “Thinking, Fast and Slow” \cite{kahneman2011thinking} in artificial intelligence \cite{booch2020thinking}, we propose a neuro-symbolic model which disentangles the visual reasoning task into neural-based perception and symbolic reasoning. We build a perception module using neural systems, similarly to System 1 (fast), which operates quickly and automatically with no effort. We employ symbolic artificial intelligence to model System 2 (slow), which focuses on cognitive abilities based on reasoning and planning.

We tackle the VideoQA CLEVRER task \cite{clevrer} which focuses on temporal and causal reasoning over physical events. The contributions of our work stand in our threefold general neuro-symbolic reasoning framework for VideoQA tasks: 1) we propose a symbolic scene representation method to combine the output of deep learning object detectors with symbolic representation, 2) we combine domain independent temporal, causal and physics background knowledge rules for reasoning on physical events in videos, 3) we present the application of our method on the CLEVRER dataset \cite{clevrer} including engineered event detection rules, outperforming state-of-the-art models on question answering accuracy.

\section{Background}
\noindent \textbf{Answer Set Programming.}
ASP is a declarative programming paradigm to solve search problems by computing answer sets of logic programs \cite{asp}. We limit ourselves to answer set programs composed of facts and normal rules of the form:
\begin{equation*}
     h\ \leftarrow\ b_1, ..., b_n,  \mbox{not } c_1, ...,  \mbox{not } c_m
\end{equation*}
where $h, b_1, ..., b_n, c_1, ..., c_m$ are all logical atoms. $h$ is described as the head of the rule, whereas the conjunction of literals (atoms and negated atoms) on the right hand side form the body of the rule. If all the literals in the body of a rule hold in the answer set, then the head must hold as well. A rule without a body is known as a fact. The negation “not” describes the negation-as-failure, where “not $c_i$” holds when all attempts to prove $c_i$ fail. Predicates will be referred to with lowercase letters, whilst variables will be in uppercase. For instance we set the $number$ predicate to hold for all integers, and use the variable $E$ to define events.

\noindent \textbf{The Event Calculus.}
The Event Calculus framework provides a logic-based formalism which inherently depicts causal relations from an event with its effects \cite{event_calculus_sha}. It has successfully been used for reasoning on and detecting events \cite{ilasp_3,oled}.

The ontology of the Event Calculus consists of: 1) actions,  temporal entities which happen over a time point, 2) fluents, quantities whose value is subject to change over time, and, 3) time points, numerical time references. Basic predicates of the Event Calculus which allow to reason over time, presented in Table \ref{tab:event_calculus} are: $initiates$, $terminates$, $happens$, $holdsAt$, $initially$, and $clipped$ \cite{event_calculus_sha}.

\begin{table}[t]
\centering
\caption{Basic Predicates of the Event Calculus framework}\smallskip
\small
\scalebox{0.95}{
\begin{tabular}{ll}
\hline
Predicates       & Meaning                                  \\
\hline
$initiates(E,\ F)$  & event $E$ causes fluent $F$ to start holding \\
$terminates(E,\ F)$ & event $E$ causes fluent $F$ to cease holding \\
$happens(E,\ T)$    & event $E$ occurs at time $T$                 \\
$holdsAt(F,\ T)$    & fluent $F$ holds at time $T$                 \\
$initially(F)$     & fluent $F$ holds from $T=0$                \\
$clipped(F,\ T)$    & fluent $F$ is terminated at time $T$
\normalsize
\end{tabular}
}
\label{tab:event_calculus}
\end{table}

We provide in Table \ref{tab:ec_rules} the Event Calculus axioms used in our framework, which allow to perform temporal reasoning. Axiom \ref{eq:initially} implies that a fluent $F$ is true as time $T=0$. Axiom \ref{eq:clipped} states that a fluent $F$ is clipped at time $T$ when an event $E$ that terminates it occurs at time $T$. In conjunction to this rule, Axiom \ref{eq:happen} affirms that a fluent $F$ is true at time $T+1$ if an event $E$ happens which causes the fluent $F$ to become true whilst no event which terminates it happens. Finally, Axiom \ref{eq:permanence} represents the law of inertia and the permanence of fluents: if a fluent $F$ holds at time $T$ and no event $E$ happens which terminates it, then it still holds at time $T+1$.

\begin{table}[t]
\centering
\caption{Event Calculus Axioms in the ASP framework}\smallskip
\scalebox{0.93}{
\begin{tabular}{l}
\hline
\small
\vbox{
\begin{align}
    holdsAt(F,\ 0) \leftarrow\ & initially(F). \label{eq:initially} \\
    clipped(F,\ T) \leftarrow\ & happens(E,\ T),\ terminates(E,\ F),\nonumber \\
    & time(T). \label{eq:clipped} \\
    holdsAt(F,\ T+1) \leftarrow\ & happens(E,\ T),\ initiates(E,\ F), \nonumber  \\
    & not\ clipped(F,\ T), \ time(T). \label{eq:happen} \\
    holdsAt(F,\ T+1) \leftarrow\ & holdsAt(F, T),\ not\ clipped(F,\ T), \nonumber \\
    & time(T). \label{eq:permanence}
\end{align}
}
\normalsize
\end{tabular}
}
\label{tab:ec_rules}
\end{table}

\begin{figure*}[t]
\centering
\includegraphics[width=0.8\textwidth]{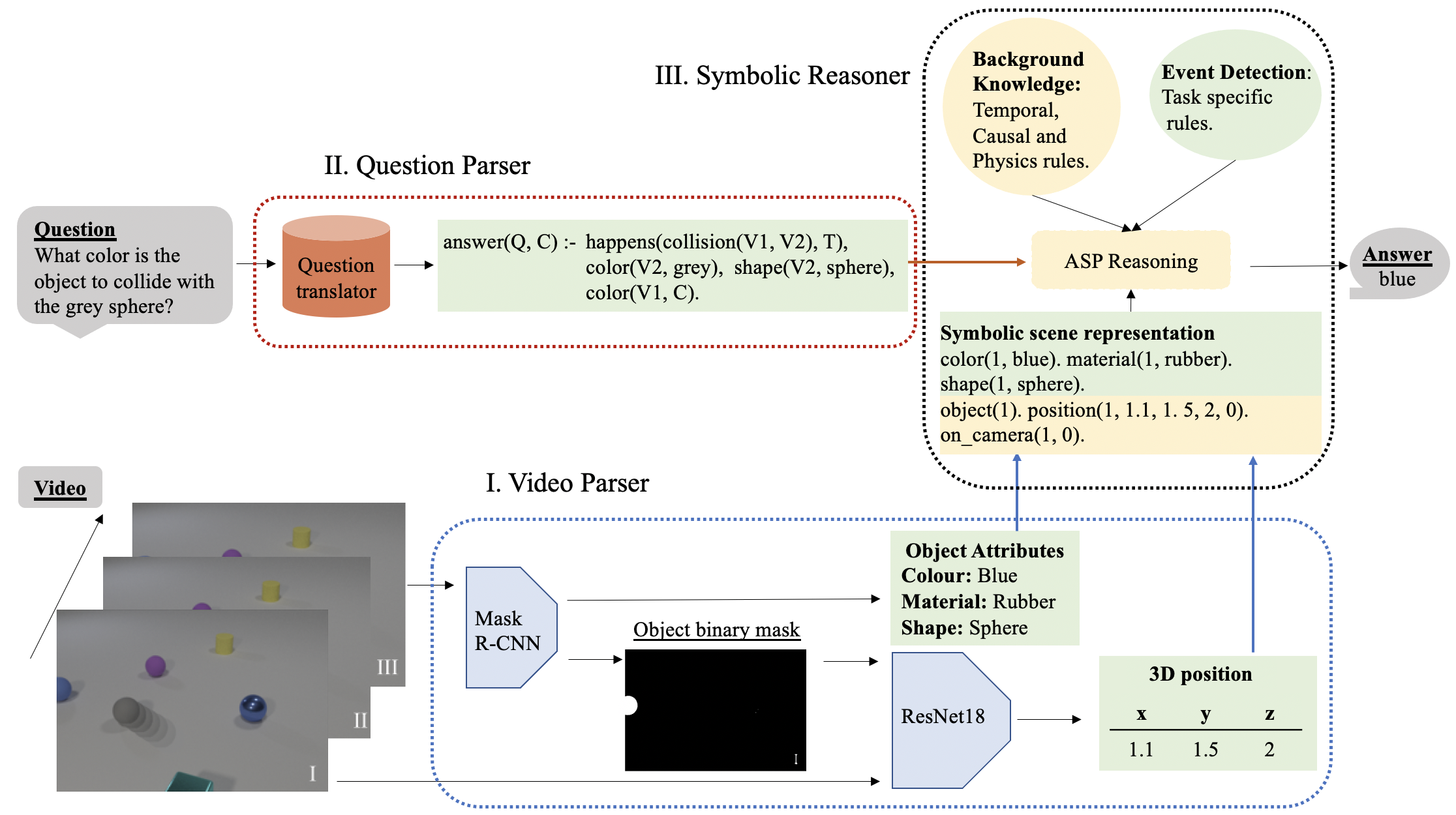}
\caption{Overview of the HySTER architecture. The video is fed to the video parser through the Mask R-CNN which segments the objects on the scene and extracts their classes. The frame is then fed alongside one of the detected object's binary mask into the ResNet18 to predict the 3D coordinate positions of the object in the scene. The information is then grounded in symbols to construct the symbolic scene representation. The natural language question is translated into a logic query and then combined with the background knowledge, the symbolic scene representation and the rules for event definition into an ASP program for the clingo ASP solver to reason over, providing an answer to the question. The yellow shaded parts refer to general components transferable between tasks. The green shaded parts refer to task specific symbolic components and information.}
\label{fig:architecture}
\end{figure*}

\section{Related Work}
Our work resides at the intersection of neuro-symbolic VideoQA and logic-based temporal reasoning on events.

\noindent \textbf{Neuro-symbolic VideoQA.} The CLEVRER dataset \cite{clevrer} serves as a reference dataset to evaluate neuro-symbolic models on complex reasoning tasks. The Neuro-Symbolic Dynamic Reasoner developed alongside the paper introduces a solution based on four pillars: 1) Video Parser, 2) Dynamics Predictor, 3) Question Parser, 4) Program Executor. The authors had proposed a similar methodology for a neuro-symbolic VQA \cite{ns_vqa}. This method allows for interpretable reasoning steps but does not set the ground for symbolic-rule learning systems which could enable to remove the human engineering from their heuristic filters and functional programs.

\noindent \textbf{Logic program in VQA.}
First-order-logic (FOL) and ASP have been used in VQA \cite{fol_vqa,asp_vqa,explicit_reasoning,gokhale2020vqalol}. A fully differentiable FOL formalism is presented in \cite{fol_vqa} where the filter sequences are translated into FOL. Doing so generalizes the filters to any domain specific language representable by FOL but does not provide insight on the decision steps taken by the model to answer the questions, which remain under the form of neural networks. The use of ASP is presented in \cite{asp_vqa} to solve the VQA task after having translated the questions into logic queries and reconstructed the scene as a logic program. This method allows for explainable reasoning and building a proof tree to justify the answer provided to the logic query. These methods have not been applied to VideoQA nor temporal and causal reasoning yet.

\noindent \textbf{Logic-based Event Detection.} Recently, hybrid methods have emerged to perform reasoning on events from videos. \cite{prapas_eve_oled} extract features from the video using a 3D CNN for an SVM classifier to detect simple events. For more complex event they apply the Event Calculus on-top of simple events detected. By using an SVM classifier to recognise events, the authors base their recognition on features which are not interpretable. In contrast, \cite{soccer_events} employ a functional program to detect simple events on symbols before applying the Event Calculus to detect more complex events. These methods do not build their event detection directly from the symbolic scene representation and general background knowledge, which hinders their generalisation across tasks.

\section{HySTER}

\begin{table*}[t]
\centering
\caption{Physics Background Knowledge Rules}\smallskip
\scalebox{0.95}{
\begin{tabular}{l}
\hline
\small
\vbox{
\begin{align}
    displacement(D,\ V1,\ T1,\ T2)\ \leftarrow\
                & position(V1,\ X1,\ Y1,\ Z1,\ T1),\ position(V1,\ X2,\ Y2,\ Z2,\ T2), \nonumber \\
                & D = (X1-X2)^2 + (Y1-Y2)^2 + (Z1-Z2)^2,\  next\_time(T1,\ T2). \label{eq:disp} \\
    disp\_greater(D1,\ V1,\ T1,\ T2)\ \leftarrow\ & displacement(D2,\ V1,\ T1,\ T2),\ D2 > D1,\ number(D1). \label{eq:disp_greater} \\
    disp\_smaller(D1,\ V1,\ T1,\ T2)\ \leftarrow\ & displacement(D2,\ V1,\ T1,\ T2),\ D2 <= D1,\ number(D1). \label{eq:disp_smaller} \\
    euc\_distance(D,\ V1,\ V2,\ T)\ \leftarrow\ & position(V1,\ X1,\ Y1,\ Z1,\ T),\ position(V2,\ X2,\ Y2,\ Z2,\ T), \nonumber \\ & (X1-X2)^2 + (Y1-Y2)^2 + (Z1-Z2)^2 <= D,\ number(D),\ V1\ !=\ V2. \label{eq:euc_distance} \\
    velocity\_change(D,\ V1,\ T2)\ \leftarrow\ & displacement(D1,\ V1,\ T1,\ T2),\ displacement(D2,\ V1,\ T2,\ T3),\ \nonumber \\
    & D >= |D1 - D2|,\ number(D). \label{eq:vel_change}
\end{align}
}
\normalsize
\end{tabular}
}
\label{tab:bckg_knowledge}
\end{table*}

In this section, we present our Hybrid Spatio-Temporal Event Reasoner (HySTER) whose implementation for the CLEVRER dataset \cite{clevrer} is shown in Figure \ref{fig:architecture}. Our model functions by combining three main components, 1) Video Parser, 2) Question Parser, 3) Symbolic Representation and Reasoning.

\subsection{Video Parser}
We define a video parser as a neural object detector which classifies objects and detects their positions in the scene, either in 2D pixel or 3D space position. The detector must provide the spatial information for each of the objects in the scene at each time frame of the video.

\subsection{Question Parser}
We define the question parser as a component which intakes a natural language question and translates it into a logic query based on the dataset vocabulary. The parser can be constructed with neural \cite{fol_vqa,singh2020exploring} or symbolic \cite{asp_vqa} natural language processing tools.

\subsection{Symbolic Representation and Reasoning}
We then construct the reasoning module pipeline which ties the outputs of the video parser and the question parser to provide an answer to the questions based on spatio-temporal properties of the video.

\subsubsection{Symbolic Scene Reconstruction}
We extract general spatio-temporal properties from any video parser through three domain independent predicates: 1) $object$ as an object identifier, 2)  $position$, providing spatial properties of the object at a given time, and 3) $on\_camera$ encapsulating the temporal information of an object being in the scene at a given time frame. We assume the existence of predicates $p_1,...,p_n$ that define task specific information extracted from the video parser such as an object's class.

\subsubsection{Background Knowledge}
We equip the model with domain agnostic background knowledge on physical definitions under the form of logic rules, built on top of the symbolic scene representation and presented in Table \ref{tab:bckg_knowledge}. The $next\_time$ and $number$ predicates hold for successive time frames and integers respectively. These rules define general physical concepts such as the displacement of an object $V1$ in the scene (Rule (\ref{eq:disp})), the euclidean distance between two objects $V1$ and $V2$ (Rule (\ref{eq:euc_distance})), or the linear velocity change of an object between two time frames (Rule (\ref{eq:vel_change})). We can apply such rules to any task which requires reasoning over the dynamics of detected objects in a scene in an explainable manner.

\subsubsection{Event Detection and Reasoning}
We employ the general physics rules from the background knowledge to construct task specific event detection rules. These rules enable the user to specifically update each event detection, which provides a direct framework for improving specific abilities of the model. This contrasts with fully retraining end-to-end neural networks and allows to target the efforts where the model fails whilst reducing the amount of data used. For instance, by detecting that the model fails in all the questions involving a given type of event $E$, we can update the rule used to define the event to improve its detection.

Temporal reasoning results from the Event Calculus framework. The symbolic representation presents the advantage of dealing with the time component $T$ directly, which allows reasoning over time with arithmetic inequalities.

We define general rules for causal reasoning on physical events over videos in Table \ref{tab:causal_reasoning}, where the events of interest must affect the state - position, velocity, acceleration - of each object it involves. Rule (\ref{eq:cause_obj}) ensures that an object related to an event is responsible for that event. Rule (\ref{eq:cause_ev}) ensures that an event causes a subsequent event if they are related. Finally, the recursive Rule (\ref{eq:cause_rec}) ensures that indirect causes are captured.

\begin{table*}[t]
\centering
\caption{HySTER Causal Reasoning Rules}\smallskip
\scalebox{0.95}{
\begin{tabular}{l}
\hline
\small
\vbox{
\begin{align}
    cause(object(V1),\ event(E1))\ \leftarrow\ & related(object(V1),\ event(E1)). \label{eq:cause_obj} \\
    cause(event(E1),\ event(E2))\ \leftarrow\ & happens(event(E1),\ T1),\ happens(event(E2),\ T2), \nonumber \\
    & related(event(E1),\ event(E2)),\ T1 < T2. \label{eq:cause_ev} \\
    cause(V1,\ V2)\ \leftarrow\ & cause(V1,\ V3),\ cause(V3,\ V2). \label{eq:cause_rec}
\end{align}
}
\normalsize
\end{tabular}
}
\label{tab:causal_reasoning}
\end{table*}

\section{Experiments}
We present the application of the HySTER framework to the CLEVRER dataset \cite{clevrer}. This VideoQA task explores temporal and causal reasoning on physical events - collision, entry, exit - taking place in synthetic videos through four question types: descriptive, explanatory, predictive and counterfactual. Motivated by temporal and causal reasoning rather than predicting the dynamics of the scene, we focus on the descriptive and explanatory questions. We then explore how we can apply our framework to the predictive and counterfactual questions.

\subsection{Video Parser}
We process each image frame in two steps: 1) object detection, 2) 3D position mapping. \cite{clevrer} released the output of their Mask R-CNN \cite{mask_rcnn} using a  ResNet-50 FPN \cite{DBLP:journals/corr/LinDGHHB16,resnets} as the backbone for object detection and scene de-rendering \cite{nsd}, which allows research to focus on the reasoning task. We use these outputs to obtain the three attributes of each detected object (color, shape, material), its segmentation mask, and a detection confidence score.

Similarly to \cite{ns_vqa}, we train a 3D mapping network on top of the object detector to provide the 3D coordinates of the objects in the scene. We assume that the 3D coordinates will provide depth information necessary to describe events in the videos using symbolic rules. We select a ResNet18 model \cite{resnets} with a 4 channel input to the network: 3 RGB for the full image, 1 Binary map for the segmentation mask. We hypothesize that the full image will provide the context required to predict the 3D position from a 2D image. We train the model using 4 random frames of 3000 videos respectively, for 5 epochs, using a learning rate of $1\times10^{-4}$ and a batch size of 8, on an NVIDIA Tesla K80 GPU. We use the mean squared error loss function applied on a vector $\boldsymbol{y} \in \mathbb{R}^3$ to learn each of the Cartesian coordinates.

\subsection{Question Parser}
We translate natural language queries into logic queries to be answered by the ASP solver applied over the symbolic reconstruction of the scene, its reasoning and event detection. We build an Extended Backus-Naur form context free grammar to build sample grammar of questions. The $\sim$150,000 descriptive questions are mapped into six different sentences, whilst the three other question types provide one sentence grammar each. The defined grammars are then mapped into logic queries through a functional program. An example of such a query is shown in Figure \ref{fig:architecture} which reveals how the model extracts the answer from the symbolic representation through the color variable $C$. The $answer(Q, C)$ predicate, where $Q$ represents the question number, holds if the body of the rule holds, extracting the queried color variable from the symbolic scene reconstruction. The questions in the CLEVRER dataset expect one word answers: $yes$ or $no$ for existence questions, a characteristic attribute for open-ended questions, an integer for counting questions. We use the clingo ASP solver \cite{DBLP:journals/corr/GebserKKS14} and logic queries to extract the attributes, the $yes$ or $no$ answer, or to count specific predicates.

\subsection{Symbolic Representation and Reasoning}
\subsubsection{Symbolic Scene Reconstruction}
Given the detected objects intrinsic characteristics and position obtained from the video parser, we build a set of ASP facts describing the scene. We recover the general predicates $object$, $on\_camera$, $position$ and additional predicates for the $shape$, $color$ and $material$ of each detected object in the scene. These predicates implicitly provide additional common sense knowledge to the model, such as the fact that \textit{cyan} and \textit{red} are colors or that \textit{rubber} and \textit{metal} are materials and allow to apply symbolic reasoning over the constituents in the scene.

\begin{table}[t]
\centering
\caption{HySTER Events and their Effects for the CLEVRER dataset \cite{clevrer}}\smallskip
\scalebox{0.93}{
\begin{tabular}{l}
\hline
\small
\vbox{
\begin{align}
    initiates(collision (V1,\ V2), \ collided (V1,& \ V2)) \nonumber \\ \leftarrow\ object(V1),& \ object(V2). \nonumber \\
    initiates(entry (V),\ present (V))\ \leftarrow\ & object(V). \nonumber\\
    initiates(entry (V),\ moving (V))\ \leftarrow\ & object(V). \nonumber\\
    initiates(move (V),\ moving (V))\ \leftarrow\ & object(V). \nonumber\\
    terminates(stop(V),\ moving(V))\ \leftarrow\ & object(V). \nonumber\\
    terminates(exit(V),\ moving(V)\ \leftarrow\ & object(V). \nonumber\\
    terminates(exit(V),\ present(V))\ \leftarrow\ & object(V). \nonumber
\end{align}
}
\normalsize
\end{tabular}
}
\label{tab:temporal_reasoning}
\end{table}

\subsubsection{Temporal Reasoning}
We define unary predicates for the $move$, $stop$, $entry$, and $exit$ events, and a binary predicate for the $collision$ event, which involves two objects. We specify the effects of detected events on the fluents to answer the questions from the CLEVRER task according to the rules presented in Table \ref{tab:temporal_reasoning} and the Event Calculus framework. These rules allow to reason on the dynamics of the scene and answer questions such as “How many objects are moving when the video ends?".

\subsubsection{Causal Reasoning}
The dataset requires to reason on the responsibilities of objects and events over the $exit$ and $collision$ events. We employ the rules defined in Table \ref{tab:causal_reasoning} by letting the $related$ predicate refer to an object taking part in the event. The objects in the CLEVRER dataset are all of similar dimensions, none of them are rooted to the ground, and no external forces other than friction apply, hence we assume that all collisions affect the dynamics of the two objects involved and respect the conditions for our rules. As such, Rule (\ref{eq:cause_ev}) ensures that a collision is responsible for a subsequent collision if one of the objects participated in both. Similar reasoning can be applied over Rules (\ref{eq:cause_obj}-\ref{eq:cause_rec}) and responsibilities for the $exit$ event.

\subsubsection{Event Detection}
We describe events occurring in the video with rules of a minimum number of predicates, which provide transparent model reasoning and can be changed to allow for model update. We present them in Table \ref{tab:event_detection}.

\begin{table*}[t]
\centering
\caption{Event Detection Rules for the CLEVRER dataset}\smallskip
\scalebox{0.95}{
\begin{tabular}{l}
\hline
\small
\vbox{
\begin{align}
    & \qquad \qquad \quad \mbox{\normalsize First Event Detection Rules} \nonumber \\
    happens(entry(V1),\ T1)\ \leftarrow\ & not\ on\_camera(V1,\ T1),\ on\_camera(V1,\ T2),\ next\_time(T1,\ T2). \label{eq:entry} \\
    happens(exit(V1),\ T1)\ \leftarrow\ & on\_camera(V1,\ T1),\ not\ on\_camera(V1,\ T2),\ next\_time(T1,\ T2). \label{eq:exit} \\
    happens(move(V1),\ T1)\ \leftarrow\ & disp\_greater(D,\ V1,\ T1,\ T2),\ next\_time(T1,\ T2),\ not\ holdsAt(moving(V1),\ T1). \label{eq:move} \\
    happens(stop(V1),\ T1)\ \leftarrow\ & disp\_smaller(D,\ V1,\ T1,\ T2),\ next\_time(T1,\ T2),\ holdsAt(moving(V1),\ T1). \label{eq:stop} \\
    happens(collision(V1,\ V2),\ T)\ \leftarrow\ & euc\_distance(D1,\ V1,\ V2,\ T),\ velocity\_change(D2,\ V1,\ T), \nonumber \\ & velocity\_change(D3,\ V2,\ T),\ not\ holsdsAt(collided(V1,\ V2),\ T). \label{eq:ev_col_v} \\ \nonumber \\
    & \qquad \qquad \qquad \qquad \mbox{\normalsize Updated Rules} \nonumber \\
    happens(entry(V1),\ T1)\ \leftarrow\ & not\ on\_camera(V1,\ T1),\ on\_camera(V1,\ T2),\ on\_camera(V1,\ T3), \nonumber \\
    & next\_time(T1,\ T2),\ next\_time(T2,\ T3). \label{eq:entry_updated} \\
    happens(move(V1),\ T1)\ \leftarrow\ & disp\_greater(D,\ V1,\ T1,\ T2),\ disp\_greater(D,\ V1,\ T2,\ T3), \nonumber \\ & next\_time(T1,\ T2),\ next\_time(T2,\ T3),\ not\ holdsAt(moving(V1),\ T1). \label{eq:move_updated}
\end{align}
}
\normalsize
\end{tabular}
}
\label{tab:event_detection}
\end{table*}

\textbf{HySTER-0.} We assume an absence of noise in the symbolic scene reconstruction which allows us to define the $entry$ and $exit$ events occurrences as Rule (\ref{eq:entry}) and Rule (\ref{eq:exit}) respectively, describing the appearance and disappearance of objects in the scene.

We build the rules for $move$ and $stop$ from the background knowledge comparative displacement Rules(\ref{eq:disp_greater}, \ref{eq:disp_smaller}) respectively. We assume that we can provide threshold displacement values indicating if an object is moving or stationary. Being in movement depends on the context and the inertia of an object, hence, we propose separate thresholds for each rules.

We define collisions in Rule (\ref{eq:ev_col_v}) by a threshold distance in conjunction with a threshold change in velocity for each involved object. The euclidean distance threshold forces the objects to be physically close, which is necessary for them to collide. We then assume a collision will lead to momentary speed changes in the two objects due to the reaction force imposed by the collision upon contact (Newton's third law of motion). In the absence of external forces and by conservation of momentum, one object would accelerate whilst the other would decelerate. Collisions between objects pairs are unique in a video, hence the presence of the $not\ holdsAt(collided(V1,\ V2),\ T)$ predicate.

\textbf{HySTER-1.} We remove our assumption of a noiseless perception module and update the HySTER-0 entry detection Rule(\ref{eq:entry}) to Rule (\ref{eq:entry_updated}), which ensures that an object must be present in the scene for two consecutive frames to $enter$ the scene. This removes noisy object detection where an object is wrongly classified over one time-step and relaxes the importance of a perfect perception module.

\textbf{HySTER-2.} On top of HySTER-1, we update the $move$ event detection rule from Rule (\ref{eq:move}) to Rule (\ref{eq:move_updated}), requiring an object's displacement to be above a threshold value $D$ for two consecutive time frames. This ensures that the object's displacement is not simply due to noise in the 3D position prediction.

\textbf{HySTER-2 (2D).} By defining a $position$ predicate with $X$ and $Y$ positions only, the model can be applied on 2D scene reconstructions from the Mask R-CNN output, without additional ground truth data to train our video parser.

We perform hyper-parameter grid searches over 500 training videos to obtain the threshold values for the rules which provide the highest question answering accuracy, making our model extremely data efficient. By updating such specific rules, we are able to target our model's weak components and improve its overall performance in the task.

\section{Results}

\begin{table}[t]
\caption{Question-answering accuracy on the descriptive and explanatory questions of the CLEVRER's test set.}\smallskip
\centering
\scalebox{0.95}{
\begin{tabular}{@{\ \ }lccc@{}}
\toprule
\multirow{2}{*}{Methods} & \multirow{2}{*}{Descriptive} & \multicolumn{2}{c}{Explanatory} \\ \cmidrule(l){3-4} & & \multicolumn{1}{l}{per opt.} & \multicolumn{1}{l}{per ques.} \\ \midrule
NS-DR                    & 88.1 & 87.6 & 79.6 \\
NS-DR (NE)               & 85.8  & 85.9  & 74.3   \\ \midrule
HySTER-2 (2D)            & 88.3 & 90.9 & 83.0 \\ \midrule
HySTER-2 (3D)            & \textbf{89.6}  & \textbf{95.9}  & \textbf{92.0} \\
\bottomrule
\end{tabular}
}
\label{tab:comparative_results_desc_expl}
\end{table}

\begin{table}[t]
\caption{Improvement pipeline accuracy on descriptive questions of the CLEVRER's test set. 0 corresponds to the HySTER with the first event detection rules. 1 corresponds to the updated event detection rule for $entry$, whilst 2 corresponds to the further update of the $move$ event detection rule.}\smallskip
\centering
\scalebox{0.95}{
\begin{tabular}{@{\ \ }lccc@{}}
\toprule
Methods - HySTER (3D) & 0 & 1 & 2 \\ \midrule
Descriptive & 86.2 & 88.8 & \textbf{89.6} \\
\bottomrule
\end{tabular}
}
\label{tab:improve_pipeline}
\end{table}

\begin{table}[t]
\caption{Question-answering accuracy on predictive and counterfactual questions of the CLEVRER's test set.}\smallskip
\centering
\scalebox{0.93}{
\begin{tabular}{@{\ \ }lclcc@{}}
\toprule
\multicolumn{1}{c}{\multirow{2}{*}{Methods}} & \multicolumn{2}{c}{Predictive}          & \multicolumn{2}{c}{Counterfactual} \\ \cmidrule(l){2-5}
\multicolumn{1}{c}{}                         & \multicolumn{1}{l}{per opt.} & per ques. & per opt.             & per ques.            \\ \midrule
NS-DR            & \textbf{82.9} & \textbf{68.7} & 74.1 & 42.2  \\
NS-DR (NE)       & 75.4 & 54.1 & 76.1 & 42.0  \\
HySTER-2 (2D)    & 79.5 & 61.5 & \textbf{79.4} & \textbf{47.1} \\
\bottomrule
\end{tabular}
}
\label{tab:comparative_results_pred_cf}
\end{table}

We compare our results with the state-of-the-art NS-DR \cite{clevrer} solution on question answering accuracy. The NS-DR outperforms other existing baseline methods such as implementations of MAC \cite{MAC} or Tbd-Net \cite{tbdnet} on the task.

In our HySTER (3D) we use the 3D position information during training whereas the authors of the NS-DR do not take advantage of this information. Hence our results are not directly comparable with theirs in this setting. However, the results prove the performance of our method, where our HySTER-2 (3D) outperforms the NS-DR by approximately 1.5, 8, and 12\% in accuracy for the descriptive, explanatory per option and per question questions respectively. We present the results in Table \ref{tab:comparative_results_desc_expl}. Even though the results from our HySTER-2 (2D) on these questions reveal that a component of the advantage comes from the use of the 3D coordinates which help in detecting collisions, these results highlight the potential of using logic rules in an ASP framework to model the reasoning task.

The 2D setting allows us to directly compare the performance of our HySTER with the one from the NS-DR \cite{clevrer} for descriptive and explanatory questions. Comparing the questions accuracy allows us to disentangle the comparison of the perception and reasoning modules, as the perception is identical. We outperform both the NS-DR and the NS-DR (NE) - which stands for “no-events” when training their propagation network and detecting collision from a heuristic rule - on these questions, as shown in Table \ref{tab:comparative_results_desc_expl}. These results reveal that our symbolic rule for detecting collision events surpasses both the propagation network collision detection ability whilst being explainable and their heuristic filter for collision detection (explanatory questions heavily rely on collision detection). This underlines the idea that apparent complex reasoning and event detection tasks can be efficiently modelled by a low number of symbolic rules and supports the implementation of ASP for the symbolic component of the model.

Table \ref{tab:improve_pipeline} shows that removing noise in the $entry$ and $move$ event predicates allows to improve the results on descriptive question accuracy. This highlights the possibility of changing one component in the model and directly seeing its impact on the overall performance. The compositional nature of our logic system inherently holds a solution to specifically notice and update failing modules.

Finally, we compare the implementation of our HySTER symbolic scene reconstruction and reasoning on top of the propagation network output from \cite{clevrer} for predictive and counterfactual questions with the NS-DR methods. The fact that we outperform the NS-DR (NE) in all question types suggests that our symbolic representation models the scene and events more efficiently. The greater performance of the NS-DR on predictive questions could be due to collisions occurring in the last frame of the videos, which are not detectable via our method due to the $velocity\_change$ predicate which compares the velocity at two time points.

\section{Analysis}

\begin{figure}[t]
\centering
\includegraphics[width=0.9\columnwidth]{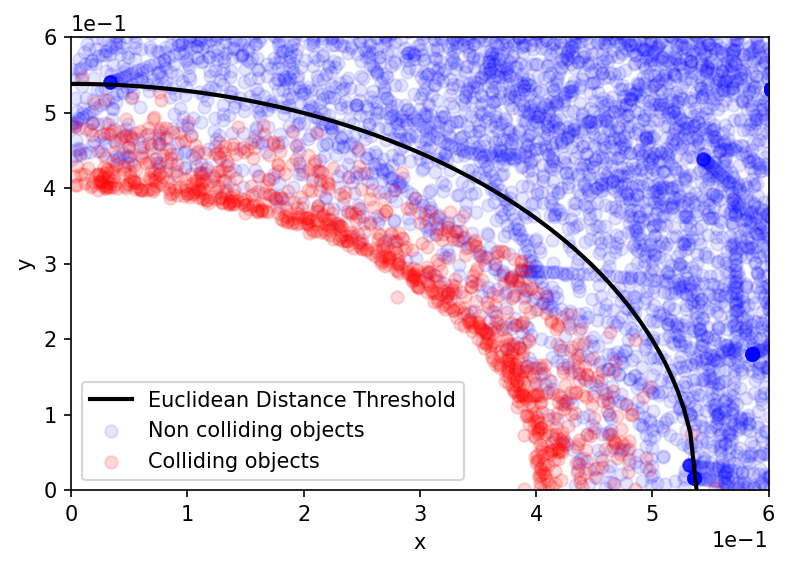}
\caption{Absolute distances in the $x$ and $y$ directions of colliding objects as red dots, and non-colliding objects as blue dots, for training videos 0-200 in the 3D setting. The optimal threshold on the euclidean distance between two objects in the velocity change rule is plotted in black.}
\label{fig:distance_threshold}
\end{figure}

\begin{figure}[t]
\centering
\includegraphics[width=0.93\columnwidth]{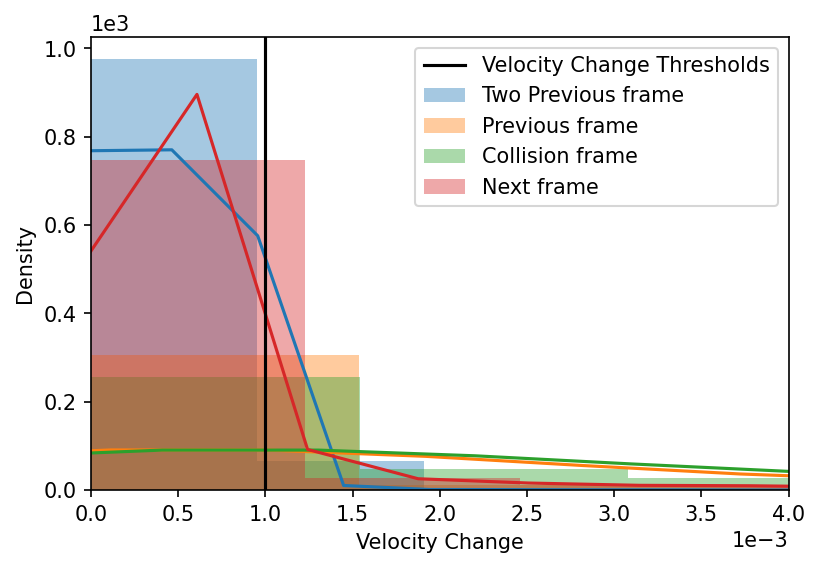}
\caption{Velocity change distributions of colliding objects: 2 frames before collision in blue, 1 frame before collision in orange, at collision in green, 1 frame after collision in red, for training videos 0-5000 in the 3D setting. The black vertical represents the thresholds from Rule (\ref{eq:ev_col_v}) (equal here).}
\label{fig:velocity_threshold}
\end{figure}

The performance of our model relies on three pillars: 1) the accuracy of the perception module, 2) the reasoning ability of the symbolic framework, 3) the event detection capability.

The accuracy of the output of the Mask R-CNN from \cite{clevrer} is crucial for efficient reconstruction of the scene into symbols. The model shows an F1 score of 0.991 (over 1000 training videos) in object detection and classification which ensures such an ability.

Symbolic AI inherently presents a framework to handle compositional, temporal and causal reasoning via symbols, the Event Calculus and causal rules. This empowers our model with human-like reasoning abilities which allows to solve some of the complex reasoning tasks from the dataset. Moreover, the logic component allows to compose with negation or counting questions, which remain difficult tasks for fully deep learning algorithms \cite{gokhale2020vqalol} and \cite{counting_deep_learning} respectively.

To understand the HySTER event detection ability we plot the collision detection thresholds over videos 0-200 from the training dataset. Figure \ref{fig:distance_threshold} shows the ground-truth distances in $x$ and $y$ direction between two colliding objects as red dots, and between two non colliding objects as blue dots. We plot the euclidean distance threshold from the HySTER-2 (3D) Rule (\ref{eq:ev_col_v}) which is represented as a circular arc: all points on that arc are at distance $D1$ form the origin, $D1$ being the square root of our computed threshold. The threshold reveals that a large proportion of non-colliding object pairs are removed, whereas most of the colliding object pairs are conserved. This prevents from having false collisions detected without hindering the detection of actual collisions.

The velocity components of Rule (\ref{eq:ev_col_v}) allows to remove non-colliding object pairs by looking into their velocity changes. We plot the distributions of velocity change of colliding objects two frames before collision in blue, one frame before collision in orange, upon collision in green, and one frame after collision in red, from training videos 0-5000 in Figure \ref{fig:velocity_threshold}. Non-colliding objects have velocity changes centered around 0, as expected by Newton's first law of motion because no force is imposed over them. The distributions of colliding objects - green and orange - are more uniform and the vertical black line representing the velocity change thresholding helps in removing object pairs which may be closer than the euclidean threshold without actually colliding, as the velocities of objects do not change. We note that the similar distributions for the previous and upon collision frames reveal that collisions can span over two time frames.

\section{Conclusion \& Future Work}
We present a model which possesses the advantages of both deep learning and symbolic artificial intelligence, yielding state-of-the-art performance on the CLEVRER dataset VideoQA task. By equipping our model with commonsense knowledge based on temporal, causal and physics rules we propose a general framework applicable for reasoning over video's physical events. These rules contain enough information to model the causal and temporal reasoning of a reference dataset.

By introducing ASP into VideoQA we expand the breadth of the symbolic component of neuro-symbolic methods and pave the way for the incorporation of research fields such as Inductive Logic Programming (ILP). Future works will employ symbolic-rule learning systems to remove the specific human engineered rules built to detect the events over the videos, whilst remaining explainable. This would also allow for automating the improvement pipeline presented. Systems such as ILASP \cite{law2020ilasp} or FastLAS \cite{fastlas} provide the ASP framework for ILP learning tasks and would enable the model to build an interpretable cognitive model of the events in the scene from videos, perception, and background knowledge directly.

\bibliographystyle{aaai}
\bibliography{references}

\end{document}